\documentclass{article}
\usepackage[utf8]{inputenc}
\usepackage[english]{babel}

\usepackage[
    backend=bibtex,
    style=numeric,
    sorting=nyt,
    natbib=true,
    maxbibnames=99
]{biblatex}
\usepackage{amssymb}
\usepackage{rotating}

\PassOptionsToPackage{hyphens}{url}
\usepackage{hyperref}
\usepackage[hyphenbreaks]{breakurl}

\addto\extrasenglish{

}

\usepackage{framed,xcolor}
\usepackage[normalem]{ulem}

\newcommand{\COe}{$\mathit{CO}_2e$}
\newcommand{\CO}{$\mathit{CO}_2$}

\usepackage{authblk}

\title{Towards Green Automated Machine Learning: Status Quo and Future Directions\thanks{This paper is published at the Journal of Artificial Intelligence Research \citep{tornede-jair23a}. The final authenticated version is available online at
\url{https://doi.org/10.1613/jair.1.14340}.}}

\author[1]{Tanja Tornede\thanks{tanja.tornede@upb.de}}
\author[1]{Alexander Tornede\thanks{alexander.tornede@upb.de}}
\author[1]{Jonas Hanselle\thanks{jonas.hanselle@upb.de}}
\author[2]{Felix Mohr\thanks{felix.mohr@unisabana.edu.co}}
\author[3]{Marcel Wever\thanks{marcel.wever@lmu.de}}
\author[3]{Eyke H{\"u}llermeier\thanks{eyke@ifi.lmu.de}}

\affil[1]{Department of Computer Science, Paderborn University, Germany}
\affil[2]{Universidad de La Sabana, Chia, Cundinamarca, Colombia}
\affil[3]{Institut of Informatics, University of Munich, Germany}

\date{June 13, 2022}

\begin{document}

\maketitle

\begin{abstract}
Automated machine learning (AutoML) strives for the automatic configuration of machine learning algorithms and their composition into an overall (software) solution \,---\, a machine learning pipeline \,---\, tailored to the learning task (dataset) at hand. Over the last decade, AutoML has developed into an independent research field with hundreds of contributions.
At the same time, AutoML is being criticized for its high resource consumption as many approaches rely on the (costly) evaluation of many machine learning pipelines, as well as the expensive large-scale experiments across many datasets and approaches.
In the spirit of recent work on Green AI, this paper proposes Green AutoML, a paradigm to make the whole AutoML process more environmentally friendly. Therefore, we first elaborate on how to quantify the environmental footprint of an AutoML tool. Afterward, different strategies on how to design and benchmark an AutoML tool w.r.t. their ``greenness'', i.e., sustainability, are summarized. Finally, we elaborate on how to be transparent about the environmental footprint and what kind of research incentives could direct the community in a more sustainable AutoML research direction. As part of this, we propose a sustainability checklist to be attached to every AutoML paper featuring all core aspects of Green AutoML.
\end{abstract}

\section{Introduction}
A machine learning (ML) pipeline is a combination of suitably configured ML algorithms into an overall (software) solution that can be applied to a specific learning task, which is typically characterized by a dataset on which a (predictive) model ought to be trained. The design of such pipelines is a time and resource consuming task due to the immense number of solutions conceivable, each of them solving the same problem but varying in performance (i.e., producing models of better or worse predictive accuracy). Therefore, aiming to find the best performing pipeline, candidates are evaluated on the dataset at hand to further adapt and improve its composition and configuration. Aiming at the automation of this process, AutoML \citep{hutter2019automated} develops methods for searching the space of ML pipelines in a systematic way, typically with a predefined timeout after which the most promising candidate is used to train the final model. The interest in the field of AutoML has rapidly increased in the recent past, and the field has broadened in scope, though predictive performance remains the main measure of interest.

In the field of AutoML, theoretical results are quite difficult to obtain, because the problem is complex and hardly amenable to theoretical analysis. Accordingly, most research contributions are of empirical nature: the proposition of a new technique or approach is accompanied by large experimental studies to show the benefits of the proposed techniques. As one is interested in techniques that work well in general, across a broad range of problems, a large number of datasets is required for evaluation. Additionally, as there is not a single AutoML system representing the state of the art (SOTA) and dominating all others, multiple competitors need to be assessed. To this end, each competitor is usually re-evaluated on the same datasets and ideally under identical conditions, for example, the same search space and hardware restrictions. Combined with long evaluation times of a single solution candidate of up to several hours or even days, like in the case of neural architecture search (NAS) \citep{elskenMH19}, all this culminates in a field that is quite resource-intensive and therefore producing immense carbon emission \,---\, much of which could be mostly avoided as we discuss later on. Note that carbon emissions or \COe{} refer to \CO{} equivalents, i.e., the amount of \CO{} with the same global warming potential as the actual gas emitted.

To fully cover the carbon emissions produced through a single published AutoML paper, one has to consider the whole production chain, starting with the generation and storage of data, the computational effort and memory needed during the development of the according AutoML system and also the final benchmark. Usually, most of the carbon emission is caused by the AutoML process, namely the evaluations of ML pipelines and the storage of the intermediate results of the search process, which is why we mostly focus on this part throughout the paper. 

However, we would like to stress that papers with a large environmental footprint are not necessarily to be doomed (this will be discussed in more detail later on). What is important, instead, is to carefully trade-off cost versus benefit. For example, a paper investing many resources in creating a benchmark, which then allows other researchers to save resources on their papers, might offer a much larger benefit than a paper investing the same resources in evaluating a method that only grants a $0.001\%$ improvement compared to SOTA methods.

The general problem is not exclusive to AutoML but similarly applies to many other AI-related fields, in particular deep learning due to its ever-increasing architectures \citep{benderGMS21,patterson2022carbon}. Lately, there has been a growing interest in being more environmentally friendly in the whole field of AI.
While AI \textit{for} sustainability is quite commonly known, sustainability \textit{of} AI has long been less of a concern. In 2019, \citet{schwartz2019green} introduced the notion of Green AI, advocating to consider the energy efficiency of AI algorithms during their development. Additionally, they propose to attach relevant information, such as the elapsed runtime of all experiments or carbon emissions to every AI paper published.
Similarly, \citet{van2021sustainable} campaigns for more work on sustainability \textit{of} AI, especially economic, social, and environmental sustainability. Even large tech companies such as Google advocate for the consideration of the \COe{} footprint as part of papers and corresponding transparency about it \citep{patterson2022carbon}.

In the spirit of the recent work on Green AI, we seek to transfer the ideas and problems to the field of AutoML, we dub this paradigm \textit{Green AutoML}. The foundation thereof is the quantification of the environmental footprint of AutoML approaches (Section~\ref{sec:measuring_environmental_impact}). In general, we identify four categories of actions the community may take towards the reduction of the environmental footprint of research on AutoML. The design (Section~\ref{sec:approach_design}) and benchmarking (Section~\ref{sec:benchmarking}) of AutoML systems are both key aspects to consider for improving the environmental footprint. Furthermore, being transparent about that footprint (Section~\ref{sec:transparency}) can give valuable additional information about an AutoML approach. Lastly, appropriate research incentives (Section~\ref{sec:research_incentives}) could direct the AutoML research in a more sustainable direction. Additionally, we elaborate on the trade-off between focusing on environmental impact and the freedom of research (Section~\ref{sec:trade-off_freedom_environmental_impact}) and on the prospects of AutoML (Section~\ref{sec:prospects-of-automl}), prior to concluding this paper.

\section{Quantifying Sustainability} \label{sec:measuring_environmental_impact}
Quantifying the sustainability of an AutoML approach is usually the first step towards a more environmentally friendly approach. However, measuring sustainability in terms of a single number, suitable as a basis for comparing different approaches with each other, is a challenging problem. 

\subsection{Measures for Post-Hoc Analysis}
\label{subsec:post-hoc-footprint-analysis}
First and foremost, it is important to differentiate between the \textit{efficiency} of an approach and the \textit{environmental footprint} of a specific experiment featuring the approach. When assessing several measures in the following regarding their ability to quantify efficiency, we always refer to the performance curve of the approach (cf. Section~\ref{sec:ecological-performance-profiles}). This curve describes the dependency between the measure of interest (e.g. runtime) and the performance (e.g. accuracy) achieved by the approach after a certain amount of budget w.r.t. that measure has been consumed. For example, we would consider a tool achieving an accuracy of $0.8$ after $30$ seconds of runtime to be more efficient than a tool requiring $60$ seconds, if runtime was a suitable measure for quantifying efficiency. Moreover, measures to quantify efficiency should be hardware-independent, very much like the O-notation for characterizing complexity in theoretical computer science, but of course not only asymptotic, because ``constants'' clearly matter in this case. 

In practice, however, approaches are often run on varying hardware with different properties, such as energy consumption. 
Furthermore, experiments often vary in terms of scope. Consider, for example, two works suggesting different approaches where one evaluates on only two datasets using an old laptop with a CPU having a high energy consumption and the other one evaluates on 100 datasets on much more efficient and also suitable hardware provided by a compute center. The second work will most likely have a larger environmental footprint because more energy is consumed than the first one, although it could potentially be a much more efficient approach. Moreover, while any reasonable measure for efficiency should be hardware-independent, measures for the environmental impact of a specific experiment on a specific machine must be hardware-dependent.

Accordingly, Green AutoML research can neither be quantified solely on the basis of the efficiency of an approach nor the environmental impact of a specific (set of) experiment(s). Instead, the two must be considered jointly. 
In the following, we discuss several such measures w.r.t.~their suitability for both measuring energy efficiency and environmental impact, which is visually summarized in Table~\ref{tab:measures}.
For the following discussion, we focus mostly on aspects of sustainability induced by computation and largely ignore other aspects, such as the software lifecycle or reliability, which are considered in general research on software sustainability \citep{calero2013systematic,lago2015framing,calero20195ws,lannelongue2021green}. In particular, the measures we discuss in the following can be attributed to the area of performance efficiency proposed by \citet{calero2013systematic} and, as the majority of the measures in that area, our discussed measures focus on estimating time or resource consumption in one way or the other.

\begin{table}[t]
    \centering
    \begin{tabular}{l||c|c|c||c|c}
        & \multicolumn{3}{c||}{Properties} & \multicolumn{2}{c}{Quantification} \\ \hline
        Measure & \rotatebox[origin=l]{90}{Hardware Independence} & \rotatebox[origin=l]{90}{Measurability} & \rotatebox[origin=l]{90}{Human Interpretability} & \rotatebox[origin=l]{90}{Efficiency} & \rotatebox[origin=l]{90}{Environmental Footprint} \\ \hline
        Runtime                     & \texttimes & \checkmark & \texttimes & \texttimes & \textopenbullet \\
        CPU/GPU Hours               & \texttimes & \checkmark & \texttimes & \texttimes & \textopenbullet \\
        Floating Point Operations   & \texttimes & \checkmark & \texttimes & \texttimes & \texttimes \\
        Energy Consumption          & \texttimes & \texttimes & \checkmark & \texttimes & \checkmark \\
        \COe                      & \texttimes & \texttimes & \checkmark & \texttimes & \checkmark \\ \hline
    \end{tabular}
    \caption{Summary of discussed measures to quantify efficiency and environmental impact, together with their properties (fulfilled: \checkmark; partially fulfilled: \textopenbullet; not fulfilled: \texttimes).}
    \label{tab:measures}
\end{table}

\subsubsection{Runtime}
\label{subsec:runtime}
Although runtime completely ignores aspects such as memory consumption, it strongly correlates with the energy consumption of the corresponding experiment, which is usually a linear function of runtime and energy consumption of the components.
However, runtime by itself is not easy to interpret by a human with regard to the size of an environmental footprint, as it is not a number that can be directly compared to something commonly used, for example the footprint of a flight. 

Nevertheless, if enough additional information, such as the energy consumption of the used hardware (per time unit) and the composition of the energy mix, are available, it is a suitable foundation for computing an estimate on the overall \COe{} footprint of the experiment at the location and time it was performed. Furthermore, compared to other measures (to be discussed in the following), runtime is rather straightforward to measure on most hardware.

Overall, runtime is a poor measure of efficiency, as it is not hardware-independent, but it is quite practical as a proxy of the environmental impact.

\subsubsection{CPU/GPU Hours}
Similarly, measuring CPU/GPU hours is both practical and easy to quantify environmental impact. Unfortunately, CPU/GPU time is often used ambiguously, as one can measure either wall clock CPU/GPU time or true CPU/GPU time, which results in different interpretations for quantifying environmental impact. On one hand, if wall clock time is used, the impact of other operations like memory access is implicitly included. This also includes overloaded main memory where the computer starts swapping. On the other hand, if real CPU/GPU time is used, those operations are partially ignored. Similarly, as before, CPU/GPU hours make it hard for a human to quantify the impact in comparison to other causes known from daily life. Moreover, counting CPU/GPU hours is a poor proxy for efficiency, as it is hardware-dependent. Nevertheless, as of now, it is one of the most practical proxies as it is easy to measure and rather easy to convert into \COe{} assuming that the CPU/GPU consistently uses a certain amount of energy and that the energy mix is known.

\subsubsection{Floating Point Operations}
Although counter-intuitive, floating point operations (FPO) are a hardware-dependent measure and as such they are not well suited for quantifying efficiency either. The hardware dependence is caused by the optimization of the compiler when the corresponding code is compiled. Depending on the degree of optimization (and the hardware the code is optimized for), the amount of FPOs can vary. At the same time, it is also problematic as a proxy for environmental footprint, because, similar to the previous measures, it ignores other elements such as memory usage. However, FPOs are often easy to measure, as most CPU/GPUs have corresponding counters that can be read. Once again, from a human perspective, this measure is hard to interpret and put in comparison with others.

\subsubsection{Energy Consumption}
Similar to runtime, energy consumption is not hardware independent but heavily depends on the energy efficiency of the hardware. Thus, while being a bad measure of efficiency of an approach, it is an excellent measure for quantifying the environmental footprint of a specific experiment on specific hardware, as energy, apart from the hardware itself, is the main external resource required to perform AutoML experiments. Moreover, from the amount of consumed energy, one can often reasonably approximate the actual \COe{} emissions caused by the experiment at the location and time of execution, if enough additional information (such as the energy mix) is available. Even more so, humans can often quite easily put energy consumption in perspective, as most people are aware of their own consumption at home. Unfortunately, measuring energy consumption is often difficult in practice, where HPC systems are used to perform experimental evaluations. While one can measure the energy used by a personal computer quite easily, using appropriate electrical instruments, measuring the energy consumption across several nodes of a cluster, which might even be shared with other users, can be very complicated. We refer to \citet{garcia2019estimation} for a comprehensive overview of available estimation methods and corresponding software.

\subsubsection{\CO{} Equivalents}
\COe{} is, in principle, an excellent and probably the most direct measure for quantifying the environmental footprint of an experiment, if the physical location and the time of execution are provided as well. However, \COe{} as a measure suffers even more from the problem of measurability than energy consumption, because it is not directly measurable. Instead, it can be computed based on the energy consumption and additional information on the energy mix. In practice, obtaining corresponding information is often not possible, and in fact, the energy mix might even vary depending on external effects, such as the weather, if it contains renewable energy components. Moreover, while the energy consumption of an experiment is mostly independent of the time and location of execution, the \COe{} footprint of the same experiment can vary drastically depending on these factors. Consequently, the energy mix should be noted as part of the footprint. For example, running an experiment at a compute center completely powered by renewable energy will result in no direct \COe{} emissions, whereas running the same experiment on a unit powered by energy produced from coal will result in much larger \COe{} emissions. \citet{patterson2021carbon} even demonstrate that the choice of the actual (deep learning) model, the compute center and the compute unit can influence the carbon footprint of a work by a factor of $1000$, and that simply choosing the right compute center location can result in a factor of up to $10$.

\subsection{Ignored Side Factors}
None of the measures previously discussed includes other side factors such as the resources used by the scheduler of the HPC system, or by a potential database holding the data, or the whole process of generating and transferring the data. Especially the latter is crucial, as reusing existing data (instead of generating new ones) makes a paper more environmentally friendly if one does not account for the data generation. Similarly, the share of the footprint caused by the production of the hardware itself is almost impossible to quantify, as this would require knowledge of details about both the lifetime and the future usage of the hardware in advance. Lastly, one actually needs to quantify the environmental impact in terms of multiple measures, as the production of hardware, the operation of HPC systems and other factors also impact the environment apart from \COe{} emissions, for example, through water usage. Overall, the problem of quantifying the environmental footprint of scientific research is extremely complex and may constitute a custom paper on its own. We refer the interested reader to the research area of software sustainability for more all-encompassing general work on the matter \citep{calero2013systematic,lago2015framing,calero20195ws}.

\subsection{Best Practices for Quantifying Sustainability} 
\label{subsec:practical-suggestions}
To measure the environmental footprint of a specific experiment, it is most practical to use the wall clock time based CPU/GPU hours as a proxy. This number is usually easy to capture without any timely investigations about the energy mix and the energy consumption, which, in case of HPC, might be impossible to obtain. Nevertheless, in case those additional information is easy to access, the environmental footprint itself can be estimated by multiplying the CPU/GPU hours with the energy consumption per unit and \COe{} of the energy mix. 
Note that although the energy mix might be green, the computation is still not for free. The mere usage of resources already leaves a footprint (tool wear, etc.), which is why saving any CPU/GPU hours of any energy mix, or avoiding any unnecessary computational time, will reduce the overall footprint of the research.
Overall, it is most important to be transparent about the environmental footprint, which might help interested readers to decide which method from a paper suits their needs. Therefore, attaching this information to a published paper is key towards achieving Green AutoML, which we elaborate on in Section~\ref{sec:transparency}.

When it comes to estimating the efficiency of an approach independent of the hardware, we do not have a concrete suggestion, as none of the solutions previously discussed is truly satisfying. Nevertheless, we think this is an important aspect that should be covered in future work. 

\subsection{Tooling and Software}
There exists software and tool support that can help to quantify environmental impact. One example is Carbontracker \citep{anthony2020carbontracker}, which tracks and predicts the carbon emissions consumed while training deep learning models.
Similarly, IrEne \citep{cao2021irene} predicts the energy consumption of transformer-based Natural Language Processing (NLP) models. \citet{parcollet2021energy} propose a framework to investigate carbon emissions of end-to-end automatic speech recognition (ASR). 
EnergyVis \citep{shaikh2021energyvis} is a more general tool, which is capable of tracking energy consumption for various kinds of machine learning models and provides an interactive view to compare the consumption across different locations. Unfortunately, it is limited to the USA at the moment. Similarly, the Machine Learning Emissions Calculator\footnote{\url{https://mlco2.github.io/impact}} \citep{lacoste2019quantifying} is an easy-to-use online tool for estimating the \COe{} footprint of a set of experiments. Similarly, the tool by \citet{lannelongue2021green} called GreenAlgorithms\footnote{\url{http://calculator.green-algorithms.org/}} estimates the \COe{} of any software artifact based on several quantities such as the CPU or GPU hours. Notably, they also consider the energy consumed by memory and other factors making it perhaps the most comprehensive tool. \cite{schmidt2021codecarbon} designed a python library, called CodeCarbon\footnote{\url{https://www.codecarbon.io/}}, directly allowing to measure the \COe{} footprint of your application within code. We close the tool section with \citep{garcia2019estimation}, which gives a comprehensive overview of methods and software for estimating the energy consumption of machine learning techniques.

\section{Design of AutoML Systems}\label{sec:approach_design}
Although, as mentioned earlier, the evaluation of AutoML approaches is presumably the largest source of negative environmental impact directly caused by AutoML research, it can arguably be seen as a symptom rather than a true cause. The evaluation of such systems is only as expensive as the search process employed, and especially the evaluation of solution candidates is often very resource intensive. Hence, we believe that developing methods inherently considering their environmental footprint is a key idea towards Green AutoML.

In general, we differentiate between three research directions that could be taken in this regard. First, one can focus on the development of methods that produce energy-efficient ML pipelines (Section~\ref{subsec:finding-energy-efficient-pipelines}). Second, one can zoom in on designing approaches that strive for a compromise between finding a good ML pipeline and minimizing the energy consumed by their search process itself (Section~\ref{subsec:energy-efficient-automl-methods}). Third, one can try to optimize the development process of AutoML approaches to be more energy-efficient (Section~\ref{subsec:efficiency-of-automl-approach-development}).

\subsection{Finding Energy-Efficient Pipelines}
\label{subsec:finding-energy-efficient-pipelines}
Developing methods whose final output is very energy-efficient does not necessarily reduce the environmental footprint of the AutoML system itself. However, it is a step in the right direction, as it most likely reduces the long-term footprint of the resulting pipeline, which will be used in production. Thus, quantifying the \COe{} saved by the deployed model should also be considered when quantifying the environmental footprint of the AutoML process itself. In principle, it is even conceivable to estimate the saved \COe{} by a deployed model to determine if the execution of the AutoML system is actually worth it. 
Although the pipeline resulting from an AutoML process in a research context is usually not deployed, this is, of course, the case for industrial applications of AutoML where the reduction in \COe{} footprint can indeed make a difference.

Moreover, finding energy-efficient pipelines is not only of interest from a sustainability driven point of view but also from a practical one. When trying to design pipelines for embedded systems such as sensor nodes or mobile devices, which rely on a battery that is not often recharged, energy efficiency becomes a core criterion.
Recent work started to target this issue \citep{he2018amc,wang2019energy,stamoulis2018designing}. For example, \citet{he2018amc} suggest AutoML approaches to compress and accelerate neural network models for usage on mobile devices and thus naturally reduce their energy consumption by reducing the number of FLOPS required for a pass through the network. Similarly, \citet{wang2019energy} suggest to iteratively enlarge a neural network through a splitting procedure which explicitly considers the increase in energy cost by splitting a certain neuron such that resulting networks are more energy-efficient than competitors. In this regard, it is also conceivable to obtain more energy-efficient pipelines as a result of the search process through the use of multi-objective AutoML methods such as \citep{elsken2018efficient,pfisterer2019multi,schmucker2021multi,candelieri2022fair}, initialized with both performance and energy-efficiency as target measures.

\subsection{Energy-Efficient AutoML Methods}
\label{subsec:energy-efficient-automl-methods}
The second direction is targeted at improving the environmental footprint of AutoML methods themselves, i.e., the underlying search algorithms. To this end, once again, several approaches are conceivable.

\subsubsection{Warmstarting}\label{subsec:warmstarting}
Warmstarting refers to a mechanism that integrates knowledge gained in prior executions into the current execution such that the optimization process does not start completely from scratch, without any prior information.
Clearly, the idea of warmstarting is to find good candidates \emph{early}, which hence would allow for shorter timeouts with the confidence that these are indeed (nearly) optimal.
Research on warmstarting techniques has been heavily boosted by AutoML challenges with short timeouts (cf.\ Section~\ref{sec:research_incentives}).

Typically, warmstarting is based on one or the other form of meta-learning \citep{vanshoren2018meta}.
That is, based on experiences from the past, which may or may not be associated with particular dataset properties, a recommendation is made for the current dataset.

A very simple case of warmstarting is when there is just a constant initial sequence that is followed.
For example, in ML-Plan \citep{mohr2018ml}, a fixed order of algorithms is provided, which is determined based on the overall average performance of algorithms across previous datasets.
Similarly, auto-sklearn 2.0 \citep{feurer2021autosklearn} scans a static portfolio that tries to cover different use cases.

Alternatively, the dataset properties can be considered in order to make recommendations. Although various forms of warmstarting have been suggested in the field of hyperparameter optimization \citep{swerskySA13,bardenetBKS13,yogatamaM14}, the first approach for AutoML we are aware of that applies this kind of warmstarting was auto-sklearn \citep{feurer2015autosklearn}.
Here, the dataset meta-features (such as numbers of instances, features, skewedness, etc.) are used to match them to datasets seen in the past.
Then, the pipelines that performed best on those datasets are considered with priority.
Meanwhile, warmstarting has become a standard technique for many AutoML systems \citep{lindauer2018warmstarting,perrone2017multiple,yang2019oboe,fusi2018probabilistic}.

\subsubsection{Zero-Shot AutoML}
\label{subsec:zero_shot_automl}
An extreme case of warmstarting is zero-shot AutoML, motivated by the idea of zero-shot learning \citep{xian2017zero}.
Here, the warmstarting mechanism recommends only a single candidate pipeline, and this one is adopted by the system without even evaluating it at all.

Interestingly, zero-shot AutoML was among the first approaches in the field.
For example, the Meta-Miner approach was based on recommendations obtained from an ontology over dataset and algorithm properties \citep{nguyen2011meta}.
Even though at that time the term zero-shot AutoML was not yet coined, no evaluation was involved in this recommendation.

A couple of zero-shot AutoML approaches have been proposed recently \citep{drori2019automl,sing2021privileged,mellor2021neural,lin2021zen}.
Those approaches leverage transfer \citep{torrey2010transfer} and meta-learning \citep{vanshoren2018meta} in an offline phase prior to their usage. During this phase, the approaches either use existing performance data of ML pipelines on a variety of datasets or generate such data on their own in order to learn a mapping from datasets to ML pipelines, which can then be queried more or less instantaneously during the actual usage. To enable such a learning, datasets usually need to be represented in terms of features, so-called meta-features \citep{rivolli2018characterizing}. For example, given a new dataset \citep{drori2019automl} compute meta-features based on a learned embedding of the dataset description, find the closest dataset from their offline training phase in terms of a measure defined on the meta-feature space and return the pipeline, which performed best on that dataset. 

Obviously, these methods cannot work completely without energy-consuming computations, but shift the need for computation away from the actual search phase to a prior offline phase offering two potential advantages. First, it allows one to schedule such an offline phase at times where renewable energy is readily available while the actual system can then still be used at any time in order to propose a pipeline for a given dataset. Second, it enables one to save large amounts of energy if the AutoML system is used excessively enough, such that the initial training phase requires less energy than using a standard AutoML system during the search.

Clearly, less extreme variants such as one-shot \citep{lake2011one} or few-shot \citep{wang2019few} approaches could be used as well.
In fact, recent challenges in this field have stimulated currently ongoing research (cf.\ Section~\ref{sec:research_incentives}).

\subsubsection{Avoiding Evaluations with a Timeout}
\label{subsec:avoiding_timeouting_evaluations}
Typically, AutoML systems that are based on a trial and error strategy, i.e., training and validating candidate ML pipelines on the given data, specify a timeout for the evaluation of those candidates.
The reason for such timeouts is that some candidates can be very time-consuming to evaluate, and thus impair exploration or, in extreme cases, even lead to a stall of the optimization process.
While limiting the evaluation time and terminating the assessment of candidate pipelines prematurely \,---\, if the specified time budget is exceeded \,---\, is a technical necessity, it severely affects the efficiency of such AutoML systems.
As pointed out by \citet{mohr2021predicting}, a large portion of computational resources are spent on evaluating pipelines that will be prematurely terminated due to a timeout yielding only very limited information for the search process.
In fact, often there is \emph{no} information at all, so the CPU/GPU time is literally wasted.

It is hence a natural objective to reduce the number of such events.
For this reason, \citet{mohr2021predicting} suggest to equip AutoML systems relying on executing solution candidates with a so-called safeguard, which estimates the runtime of a pipeline prior to execution and prohibits its evaluation in case a timeout is likely to occur.
Similarly, \citet{yang2019oboe} involve a runtime prediction component allowing one to maximize the information gain in comparison to the time spent on the evaluation of a pipeline.

The problem of pipeline runtime prediction is arguably harder than the one of predicting the runtime of ``atomic'' learning algorithms alone.
In fact, there has been a lot of work on algorithm runtime prediction in general (e.g., \citep{hutter2014algorithm,tornede2020run2survive,huang2010predicting,smith-MilesH11,eggensperger2020neural}).
However, as shown by \citet{mohr2021predicting}, predicting pipeline runtimes is more than just aggregating runtimes of its components, because the output of components, e.g., pre-processing steps, often impacts the runtime of subsequent components, e.g., other pre-processors or the learner.

Concepts similar to the safeguard mentioned above can be found in the domain of algorithm configuration \citep{hutter2009paramils,ansotegui2009gender,hutter2011sequential}, where an algorithm should be configured to optimize its runtime and racing or adaptive capping mechanisms \citep{hutter2009paramils} are used.
Essentially, adaptive capping prematurely terminates the evaluation of solution candidates to speed up the optimization process based on some criterion, such as bounds on the achievable performance.

\subsubsection{Multi-Fidelity Performance Measurements}
An alternative approach is to make evaluations so cheap that there is no longer a need to consider timeouts.
The idea here is to use a cheap-to-compute function to approximate the \emph{relative} performance of a candidate pipeline.
Of course, the performance of a candidate is always only estimated, but often this is done through (costly) cross-validation procedures using a lot of data.
The idea of low-fidelity estimation is to have a cheap estimator that is trained on low-cost approximations and which is faithful with respect to the \emph{ordering} of candidates.

As one approach, one could try to pick models based on evaluations using subsamples of the data for which the models are cheap to evaluate.
In fact, this approach has been proposed early on and was shown to be quite effective \citep{petrak2000fast}.
While this approach assigns a constant (prior) evaluation sample size, more recent approaches in the area of \emph{multi-fidelity optimization} add the evaluation fidelity (sample size) as a degree of freedom to the optimizer.
Typical resource candidates to influence the degree of evaluation fidelity are the size of the training set or the number of iterations for iterative learning algorithms such as gradient descent.
When being able to evaluate performance at different degrees of fidelity, adapted Bayesian optimization methods can be leveraged in order to optimize machine learning pipelines in a cost-effective manner \citep{kandasamy_multi-fidelity_2017,klein_fast_2017,falkner2018bohb,wu_practical_2019,candelieriPA21,zimmer_auto-pytorch_2021,candelieri2022fair,feurer2021autosklearn}.
In particular, the FABOLAS approach \citep{klein_fast_2017} actively trades off the sample size against the expected performance. Similarly, other optimization methods based on multi-armed bandits or differential evolution can be equipped with ideas from multi-fidelity optimization \citep{li_hyperband_2017,awad2021dehb}.

Orthogonal to this, it is possible to reduce the number of repetitions in a cross-validation.
That is, one trades the stability obtained from various validation iterations for evaluation speed.
While k-fold cross-validation is not very flexible in this regard, Monte-Carlo Cross-Validation (MCCV) can be considered in a fine-granular manner configuring \emph{both} the sample size \emph{and} the number of iterations.
To our knowledge, there are no studies that analyze let alone dynamically fine-tune MCCV in order to reduce computational time without losing the order on the candidates.

\subsubsection{Early Discarding of Unpromising Candidates} \label{subsec:early_discarding_of_unpromising_candidates}
Early discarding means to abandon a candidate early in its training process if it gets apparent that it will not be competitive.
Early discarding is operationalized via empirical learning curves.
That is, by analyzing the partially available learning curves, one can decide on the relevance of the respective candidate.

We distinguish between approaches that adopt early discarding based on a horizontal or vertical model selection strategy.
In a horizontal scenario, a portfolio of candidates is fixed in the beginning, and learning curves are grown simultaneously for increasing anchor sizes (hence horizontally from left to right).
At each anchor, a set of candidates is dropped.
This is the core idea of successive halving \citep{jamieson2016successivehalving}.
Horizontal approaches are also sometimes called \emph{multi-fidelity} optimizers.
In a vertical scenario, candidates are generated \emph{sequentially} and each of them is evaluated on increasing anchors until it can be predicted that it will not be competitive.
This approach was considered first for neural networks in \citep{domhan2015speedingup} and more recently in the Learning Curve based Cross-Validation scheme (LCCV) for all types of learners \citep{mohr2021towards}.
The authors showed that, even if all learners are non-iterative, the time required to evaluate a specific portfolio can be reduced by over 20\% on average compared to a cross-validation.
For portfolios with iterative learners, one would expect this improvement to increase even further.

Between these extremes, there are also hybrid approaches.
For example, Hyperband \citep{li_hyperband_2017} follows a horizontal approach but adds new candidates to the portfolio at each stage.
Another approach is Freeze-Thaw-Optimization \citep{swersky2014freezethawbo}, which allows for pausing training processes and resume them later if the candidate appears attractive again.

\subsubsection{Energy Consumption as Part of the Objective Function}
It is also conceivable to make the AutoML search algorithm directly aware of the energy it consumes. For example, one could adapt Bayesian optimization (BO) for AutoML \citep{thornton2013auto,feurer2015autosklearn,komer2014hyperopt} by incorporating a version of expected improvement, which considers the energy consumed when evaluating the next solution candidate. In analogy to the idea of expected improvement per second \citep{snoek2012practical}, one way of considering energy consumption in the optimization process is to employ expected improvement per kWh of consumed energy as an acquisition function, i.e., 
\begin{equation}
    EIkWh(p) = \frac{EI(p)}{kWh(p)} \, ,
\end{equation}
where $EI(p)$ denotes the expected improvement associated with pipeline $p$, and $kWh(p)$ denotes the estimated energy consumption associated with evaluating pipeline $p$, might be a good candidate for further investigation. With such an acquisition function, BO is guided towards carefully weighing between the information gain of a solution candidate and its execution cost. However, similar to the methods presented in Section~\ref{subsec:avoiding_timeouting_evaluations}, this idea requires knowledge about the energy consumption of a specific pipeline prior to its execution. First work on corresponding estimation techniques exists \citep{cao2021irene,stamoulis2018hyperpower,anthony2020carbontracker}. \citet{stamoulis2018hyperpower} also suggest a similar acquisition function explicitly incorporating energy constraints on models. However, they only consider the amount of energy required for inference of a trained network instead of the energy for training, which we are interested in. Similarly, instantiating Hyperband \citep{li_hyperband_2017} with energy as a budget is also an option. In principle, one could even consider combining both ideas in order to create an adapted version of BOHB \citep{falkner2018bohb}, which essentially constitutes a BO-Hyperband hybrid.

\subsubsection{Exploiting Heterogeneous Hardware Resources}
As another possible approach, one could consider the design of AutoML systems that exploit the heterogeneity of solution candidates w.r.t.\ their energy consumption on different hardware to improve the overall consumption.
To this end, one could exploit a heterogeneous (w.r.t.\ the types of computational devices, i.e., CPUs, GPUs, FPGAs, etc.) cluster and then schedule the evaluation of a solution candidate on the hardware best suited for the current model in terms of energy-efficiency. For example, while a neural network should be evaluated on a GPU, other learners can potentially be more efficiently executed on CPUs. This might not necessarily speed up the AutoML search process in terms of time, but can still result in an improvement regarding energy consumption.
While we believe that this is a potentially interesting line of research, we are not aware of any work in this direction.

\subsubsection{Intelligent Stopping Criteria}
\label{subsec:intelligent_stopping_criteria}
AutoML systems can also be improved by implementing intelligent stopping criteria, which consider if it can be assumed that an improvement is possible within the remaining runtime. The main idea is to decide whether the granted runtime is actually needed or if the search can be stopped early. Those considerations are similar to the ideas of early stopping \citep{prechelt1998early} in machine learning or early stopping criteria from the field of metaheuristics \citep{gendreau2010handbook}. 

As an example, the concepts proposed in LeanML \citep{samo2021leanml} could be used as such an intelligent criterion. It is based on the idea of estimating the highest achievable performance on a dataset, which can then be used to prematurely stop the AutoML search process, once the chance of finding a better solution in the remaining time is very small.

A work in this direction, which won the best paper award at the first AutoML conference in 2022, was recently presented by \citet{makarova2022automatic}, who suggest stopping hyperparameter optimization when the validation performance is presumably close to the achievable performance. To this end, they analyze the estimated difference between the validation loss and the test loss and stop when it is roughly equivalent to the estimation error associated with this difference.

\subsubsection{Use of Saved Resources}
As we have seen, there are various possibilities to save runtime, such as avoiding timeouting evaluations (cf. Section~\ref{subsec:avoiding_timeouting_evaluations}), discarding unpromising candidates early (cf.\ Section~\ref{subsec:early_discarding_of_unpromising_candidates}) or implementing intelligent stopping criteria (cf.\ Section~\ref{subsec:intelligent_stopping_criteria}).
In general, there are two ways to make use of saved resources.
One option is to terminate early, which directly influences the environmental footprint. Alternatively, the search could be continued such that more solution candidates can be considered, and eventually less of the allocated resources will be wasted.
Accordingly, it is important to note that simply reducing the amount of wasted resources does not automatically coincide with energy savings unless the overall search budget is reduced according to the search time saved.
Nevertheless, such improvements can be valuable even when the search time is not reduced as the benefit/environmental cost ratio increases.

\subsection{Efficiency of AutoML Approach Development}
\label{subsec:efficiency-of-automl-approach-development}
The third direction is to address the development phase of AutoML approaches. Due to the complexity of AutoML approaches, one usually needs multiple evaluations until the system is (mostly) free of bugs and all concepts are working as intended. Although a lot of effort can be made to decrease the environmental footprint of an AutoML approach or its final pipeline, in a research context, usually the development phase makes up a significant part of the actual environmental footprint and, therefore, the carbon emissions produced. In the following, we elaborate on a few concepts that can be applied to avoid wasting resources.

A very basic concept, which can decrease the environmental footprint drastically, is a simple caching strategy. The key idea is to cache the results of evaluated pipelines so that at least some solution candidates do not need to be reevaluated when restarting the AutoML approach. In such a case, the runtime should, of course, be cached as well in order to add it to the elapsed time, such that a rerun does not benefit from previous executions. This is especially important for AutoML settings, where the evaluation of a single solution candidate takes a lot of time, like in neural architecture search \citep{elskenMH19} or in the field of AutoML for predictive maintenance \citep{tornede2020automl,tornede2021coevolution}.

Another way to decrease the environmental footprint is to work with a very small development dataset and search space. Based on that, testing the implementation can be done within a short amount of time, thereby again decreasing the runtime and saving both development time and energy.

\section{Benchmarking}
\label{sec:benchmarking}
Since AutoML is an empirical research branch, experimentation is an inherent part of the research in general and at the heart of almost every publication. Due to the complexity of AutoML systems, theoretical results are rather hard to obtain and usually assume some simplifications, which  in turn limits the scope of conclusions that can be drawn from the results. Moreover, most theoretical results are accompanied by experiments to show that these are also reflected in practical scenarios. Typically, experimentation in AutoML not only involves experimentation with the newly proposed method but also with its competitors as baselines to demonstrate that the novel method is indeed superior to the current SOTA.

However, as in the case of the base algorithms being selected and configured by AutoML systems, the performances of AutoML systems themselves are complementary to each other \citep{mohr2018ml}. Hence, there is not a single best AutoML system representing the current SOTA, but rather an array of competitive methods. As a consequence, comparisons are carried out to a (growing) set of baselines, in turn, leading to increased computational costs. While \citet{feurer2015autosklearn} report computational cost of 11 CPU-years (CPUy), the experimental data of \citet{mohr2018ml} is the result of 52 CPUy worth of computations. In \citep{wever2021automl}, the experimental study is as extensive as 84 CPUy. In the sub-field of neural architecture search (NAS) \citep{elskenMH19}, computational costs for experimentation can even be higher. Single AutoML runs use the computational power of 450 GPUs for 7 days \citep{real2019regularized} or even 800 GPUs for 28 days \citep{zoph2017neural}\footnote{Assuming a GTX 1080 Ti with a TDP of 250W, the GPU power demand per run amounts to 134.4MWh, which is the equivalent of the yearly power consumption of roughly 30 4-persons households with a power demand of 4,250kWh each. This does not yet include the power consumption of the remaining system. On current AWS GPU Nodes, a single such run costs 483,840\$.}. Note that the published numbers usually only take into account the computational costs of the results eventually presented and not those incurred during development for testing.

The main issue causing baselines to be executed repeatedly with almost every study is that there is no gold standard for the experiment setup.
That is, datasets as well as specifics of the setup (such as hardware resources, timeouts, search spaces, configurations of the AutoML systems) vary from study to study, impeding not only comparability across publications but also reproducibility as some important details concerning the experiment setup might be missing.
To address these issues, various benchmarks have been proposed in order to foster reproducibility and comparability (Section~\ref{sec:benchmarking-reproducibility-and-comparability}) as well as sustainability (Section~\ref{sec:benchmarking-sustainability}). Moreover, we propose to consider ecological performance profiles when benchmarking AutoML systems (Section~\ref{sec:ecological-performance-profiles}).

\subsection{Reproducibility and Comparability}
\label{sec:benchmarking-reproducibility-and-comparability}

From a research perspective, benchmarks serve the purpose of providing a common platform for empirical research.
The main goal is to establish or increase the comparability and reproducibility of results.
For the research area of automated machine learning this includes, for example, the definition of certain variables such as the set of datasets to be examined, the target metric to be optimized, time bounds for both the evaluation of single candidate solutions and the total runtime of the AutoML system, the hardware to be used, the search space definition, etc.
According to the complexity of AutoML systems and the many possible configurations, there is a large number of variables that can and need to be fixed by benchmarks or explicitly left open.

For example, the OpenML \citep{vanschoren2014openml} AutoML benchmark \citep{gijsbers2019open} features 39 datasets, for each of which an AutoML system is given a total of 4 hours to search for a suitable pipeline.
Every such run is repeated ten times with different seeds.
Furthermore, the benchmark suggests to use Amazon AWS \textit{m5.2xlarge} compute nodes, featuring an Intel Xeon Platinum 8000 Series Skylake-SP processor with 8 CPU cores and 32GB memory. The reason for this is that, on the one hand, the specifications are in line with the hardware specifications of the majority of AutoML publications and, on the other hand, in principle anyone can get access to such compute nodes.
According to the specifications of the benchmark, the evaluation of a single AutoML system requires 12,480 CPUh. Hence, for estimating the computational resources of an entire study, the amount of CPU hours can simply be multiplied by the number of considered AutoML systems or baselines. Fortunately, when using the same computing infrastructure as well as the exact same experiment setup, there is no need to re-evaluate already benchmarked AutoML systems, as the results should be comparable in principle.

However, it is important to note that all the different criteria need to be met exactly in order to ensure comparability.
In the literature \citep{liu2019darts,liu2018progressive}, results are sometimes borrowed one-to-one from previous publications without accounting for changes in the experiment setup based on which the newly proposed method is evaluated. While this is certainly using as little energy as possible, the results are incomparable and valid conclusions can hardly be drawn. Obviously, the energy consumption is relatively low, however, energy efficiency in turn is poor since the information obtained through investing energy is not as valuable as desired. Various benchmarks for AutoML systems have already identified a plethora of confounding factors \citep{balaji2018benchmarking,gijsbers2019open,wever2021automl,zoller2021benchmark}, which hinder interpretation of the results and insights derived from them.

\subsection{Sustainability and Democratizing Research}
\label{sec:benchmarking-sustainability}
Beyond the properties of reproducibility and comparability, benchmarks can also serve another interest, namely to make experiments more sustainable and to democratize research on extremely computationally expensive problems. For example, several benchmarks have already been presented in the AutoML sub-field of neural architecture search (NAS), for which all possible architectures within a certain search space have been evaluated once, and the determined performances are stored in a lookup table \citep{ying2019bench,dong2020bench,zela2020bench,klyuchnikov2020bench,li2021hw}. When evaluating new approaches for optimizing neural architectures, it is sufficient to look up the performance of a solution candidate in that table instead of actually training and validating the respective architecture. Typically, the lookup tables of such benchmarks comprise the performance values of roughly 50,000 different architectures. While the costs for creating such a benchmark are obviously quite substantial, they represent a one-time investment. Moreover, since the evaluation of such a large number of architectures requires massive amounts of computations with GPUs, these benchmarks also enable researchers and practitioners, who do not have access to such computational resources, to do research on NAS.

As the number of possible architectures is quite limited in the aforementioned benchmarks, \citet{siems2020bench} propose a surrogate model  to predict the performance value of a neural architecture.
To this end, the surrogate model is trained on the performance values of roughly $60,000$ different architectures and is found to generalize these training examples quite well.
Using a surrogate model allows for more flexibility, as it can also provide performance estimates for candidate solutions that have not been evaluated at all. Hence, benchmarks centered around such a surrogate model can be even more sustainable.
However, these models need to be constructed with care to ensure that they do not mislead the research on new methods.
Furthermore, in order to reach the level of quality of the surrogate model, \citet{siems2020bench} need to create an extensive amount of training data, namely, $60,000$ architectures have been evaluated for this purpose.
While this means an immense amount of computation, it is again a one-time cost that amortizes with each subsequent evaluation.
When employing a surrogate model, however, additional constraints on how to use the benchmark need to be imposed. More specifically, the benchmark specification requires the surrogate model to be used in a query-only mode, i.e., any approach is only allowed to request performance estimates for some neural architecture. Consequently, it excludes approaches exploiting the surrogate model, e.g., by analyzing its internal structure.

All in all, benchmarks are powerful tools to make research on AutoML more sustainable. In particular, they can avoid repetitive evaluations of candidate solutions. This not only helps to save energy but also enables institutions that cannot afford the necessary resources to research on this topic. Moreover, in general, research can also be accelerated, since evaluations of candidate solutions require only milliseconds instead of minutes, hours, or even days. Consequently, the use of benchmarks should clearly be advocated and also requested, since several advantages that benchmarks bring with them can be combined in this way. However, within the community, care should also be taken to ensure that multiple very similar or possibly even identical benchmarks are not developed in parallel, as this would again unnecessarily drive up energy costs. Ideally, the development of benchmarks should be a community effort and thus be communicated at an early stage, as it was done in the case of DACBench\footnote{\url{https://github.com/automl/DACBench}}, for example.

\subsection{Ecological Performance Profiles}
\label{sec:ecological-performance-profiles}

\begin{figure}
    \centering
    \includegraphics[width=0.9\textwidth]{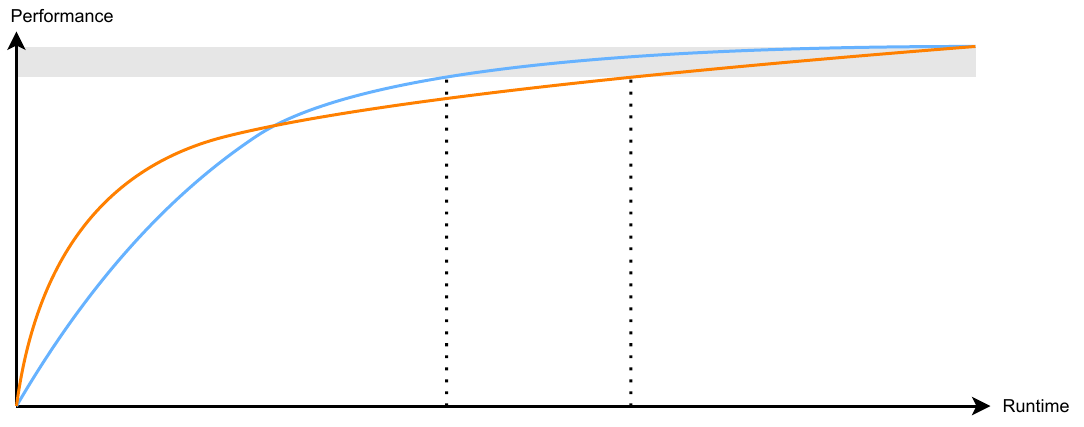}
    \caption{Example figure of two performance curves, where the blue tool achieves a good performance value (grey area) long before the orange one.}
    \label{fig:example-ecological-performance-profiles}
\end{figure}

When assessing the performance of an AutoML system, it is not sufficient to compare the final performances after the complete run has passed. Some AutoML approaches show strong anytime performance while others may yield the best final performances. Instead, one should take a look at the performance curve of the system, i.e., what solution quality can be expected after a certain amount of computational budget has been consumed. An example of the ecological performance profiles of two competing AutoML systems can be found in  Figure~\ref{fig:example-ecological-performance-profiles}, where the blue tool achieves a good performance long before the orange tool. Ideally, we would look at an ecological performance profile that relates predictive performance to the amount of \COe{} emitted for computations. However, as already discussed in Section~\ref{sec:measuring_environmental_impact}, it is more practical to investigate the CPU/GPU hours instead of the actual environmental impact as a proxy of it.
While we can assume for any reasonable AutoML system that the performance profiles are monotonic, i.e., exhibiting better performances with an increasing budget, the curves of the individual approaches may still cross each other. The ecological performance profile of an AutoML system may depend on several properties such as the runtime, the degree of parallelism, how well heterogeneous execution environments can be utilized, etc. Assuming the same hardware setting has been used for all competitors, runtime can indeed be a good proxy. Having access to such performance profiles would allow the user to choose the AutoML system which is most suitable for their budget and \COe{} footprint. A single evaluation at a fixed point is not a sufficient basis for such a decision. However, those performance profiles still depend on the hardware used for assessing the performance of the approaches and deriving their profiles.

\section{Transparency}
\label{sec:transparency}
Being transparent about the efficiency and environmental footprint is a key step towards Green AutoML and in general a more sustainable world, which we believe should be mandatory when publishing a paper. The acceptance of a paper is usually based on its novelty and performance improvement, but the environmental footprint should also be considered. One may wonder whether rejecting a paper with strong results solely for reasons of environmental impact is unreasonable \,---\, after all, the pollution cannot be undone, and pollution with published results might be better than pollution without results. Nevertheless, authors should be incentivized to avoid environmental impact right from the beginning. Accordingly, as a first step, we advocate extending checklists recently introduced in major machine learning conferences by the questions noted below. Such checklists need to be filled out prior to paper submission and are intended to help authors reflect on issues such as reproducibility, transparency, research ethics, and societal impact. Obviously, authors could also be asked for the compute resources used for their paper, whether any measures have been taken to quantify or reduce the consumption, etc.

One aspect of being transparent is to include information about the compensation of the footprint, and in the best case, also about how it is compensated. If it is done via planting trees, for example, those trees have to survive $10$ years to compensate the footprint. This might also create more awareness that it takes a long time until the consumed resources are actually compensated, and therefore researchers might run experiments with more caution.

In addition, we as a community should strongly advocate the publication of failed attempts and negative results in order to foster Green AutoML or more general green science. We all know cases, where despite a promising idea, a lot of work, and experiments, one just could not get the idea to work as well as expected in practice and hence, it was never published. This is not only unfortunate from an environmental but also from a scientific perspective. First, due to the rapidly growing scientific community (especially in AI), chances are high that someone else might work or might have worked on the same idea at a distance in time and potentially come to a similar conclusion. This work would not have been done multiple times if the negative result was published. Second, withholding negative results can decelerate scientific progress as chances are high that even if one deems an idea fully explored, someone else might have a good idea on how to turn a negative result into a positive one by making the correct adjustment to the idea. Especially, due to the large amount of computational resources needed to perform research in the field of AutoML, it is even more important to find a way to share failed attempts. A compiled list of journals targeting negative results can be found online\footnote{\url{https://www.enago.com/academy/top-10-journals-publish-negative-results/}}.

\subsection{Sustainability Checklist}
Apart from all aspects mentioned above, we believe that a key aspect towards Green AutoML is to be transparent about the environmental impact. Accordingly, authors should provide a summary of all the efforts made to make the AutoML design, development, and evaluation more sustainable, as well as information about the environmental footprint of their work, such that other people can take this information into account when deciding which method to use. We propose to attach a sustainability checklist to each paper submitted, that includes the following aspects:

\subsubsection*{Design, Development and Evaluation}
\begin{itemize}
    \item[$\square$] \textbf{What key aspects does your approach design include to be efficient?} State if and how your approach is built with efficiency in mind. For example: Do you use warmstarting? Do you apply multi-fidelity evaluations? Does it avoid timeouting evaluations? Is your approach aware of its environmental impact, like expected improvement per kWh or per g/\COe?
    \item[$\square$] \textbf{What steps did you consider during the development to reduce the footprint of the development process?} State whether and how you attempted to avoid wasting resources. 
    \item[$\square$] \textbf{Does the evaluation consider a metric related to environmental footprint?} State whether and how the evaluation does not only consider performance but also other metrics, e.g., related to improvement per invested \COe{} tons.
    \item[$\square$] \textbf{Does the work yield an improvement over SOTA in terms of efficiency or environmental impact?} State whether and how the presented method improves compared to the SOTA in terms of efficiency or environmental impact.
    \item[$\square$] \textbf{Did you add your approach to an existing benchmark?} State whether you added your approach to an appropriate existing benchmark.
    \item[$\square$] \textbf{Did you make the generated data publicly available?} State whether you created a publicly available repository of the data generated during the creation of the work such as pipeline evaluations on datasets.
\end{itemize}

\subsubsection*{Resource Consumptions}
\begin{itemize}
    \item[$\square$] \textbf{What resources have you used for the final evaluation?} State the type of CPU/GPU hours and the kind of parallelization that has been used.
    \item[$\square$] \textbf{How many CPU/GPU hours have been used for the final evaluation?} State the amount of CPU hours that have been used for the evaluation presented in the paper.
    \item[$\square$] \textbf{What is the used energy mix?} State if and to what degree your experiments were run using renewable energy. To this end, reaching out to your compute center provider will most likely be necessary.
    \item[$\square$] \textbf{What is your footprint?} State how many tons of \COe{} you produced during the creation of the paper.
    \item[$\square$] \textbf{Did you compensate the carbon emissions?} Especially if non-renewable energy has been used for the creation of your paper, state to what degree the corresponding amount of carbon emissions have been compensated, e.g., through supporting a carbon offset project. 
\end{itemize}

\section{Research Incentives}
\label{sec:research_incentives}
In order to foster research on Green AutoML, we believe that it is of great help to set according research incentives. 

A potential driver for advances in Green AutoML are scientific challenges.
Those challenges often impose narrow and rigidly realized timeouts, which oblige approaches not only to adhere to the timeouts but also to be \emph{fast} in finding good solutions.
For example, the AutoML challenge realized in 2015-2018 granted only 20 minutes of compute time to identify a model \citep{guyon2019automlchallenge}.
In some cases, datasets assumed sizes of several hundreds of megabytes, in some cases over a gigabyte.
In such situations it is impossible to conduct any sort of ``exhaustive'' search, so very resource efficient approaches needed to be developed; maybe even entirely abstaining from search in a classical sense.
One immediate evidence for the impact of challenges is the PoSH \citep{feurer2018practical} approach, which combines portfolios with successive halving to be successful in the previously mentioned challenge, and which is the basis for auto-sklearn 2.0 \citep{feurer2021autosklearn}.
This challenge has been continued for the problem of deep learning \citep{el2021metadl} and in its latest edition, which is active at the time of writing, looks at the problem of few-shot deep learning.
Since challenges at a time provide meaningful baselines, they are a valuable resource for researchers in supporting the efficiency of their approaches and hence serve as catalysts for efficient approaches in whole research fields.

Moreover, funding agencies can advocate research on Green AutoML, or more broadly, green/sustainable AI, by a variety of means. For example, they can complement the proposal evaluation criteria by considering both the potential environmental footprint and the environmental benefits of a project (proposal). To this end, applicants need to include this information in their proposals, of course. Similarly, special programs can be initiated especially targeted at sustainable and green AI. As a commendable example, the German Research Foundation (DFG) has recently released a press release\footnote{\url{https://www.dfg.de/en/service/press/press_releases/2020/press_release_no_38/index.html}} promoting their focus on sustainability. To this end, they have set up a special committee (called the German Committee Future Earth) targeted at advancing interdisciplinary research on sustainability. Furthermore, since the end of 2020, carbon emissions resulting from business trips as part of DFG funded projects can be compensated for by buying according compensation certificates, which can be accounted as travel costs\footnote{\url{https://www.dfg.de/en/service/press/press_releases/2020/press_release_no_59/index.html}}. In fact, new proposals can even contain a special category for \COe{} compensation as part of the travel expenses category. Unfortunately, according to personal communication with the DFG, it is currently not possible to budget money for \COe{} compensation of experiments.

Finally, journals and conferences can advocate special issues and special tracks focusing on green AutoML, or, more broadly green/sustainable AI. Similarly, special awards for work on sustainability are tools to put a spotlight on noteworthy work, but also the topic itself. As an example of such efforts, in 2013 the AAAI offered a special track on computational sustainability and AI\footnote{\url{https://www.aaai.org/Conferences/AAAI/2013/aaai13csaicall.php}}. However, the track was mainly focused on applying AI \textit{for} sustainability in general and less on improving the sustainability \textit{of} AI itself. Remarkably, in 2021 the first conference on Sustainable AI\footnote{\url{https://www.sustainable-ai.eu/}} was held in Germany organized by Prof. Dr. van Wynsberghe from the University of Bonn. The first AutoML conference\footnote{https://automl.cc/} held in 2022 featured parts of the checklist suggested by us in an earlier version of this manuscript in their submission form.

\section{Discussion on Trade-Off between Freedom of Research and Environmental Impact}
\label{sec:trade-off_freedom_environmental_impact}
So far, we have discussed why we believe that sustainability and, in particular, the environmental footprint induced by (research on) AutoML is an important topic, which is already addressed in some works, but should be focused on much more. However, an important question is to what degree this should be done and to what extent this limits the potential of the freedom of research. For example, questions of the following nature arise:
\begin{itemize}
    \item How strong should incentives made by a conference or a funding agency be? 
    \item What is considered a wasteful or too extensive evaluation in a paper? 
    \item When is a certain improvement, e.g., in terms of performance or another measure, worth the invested resources?
    \item What is a reasonable degree of transparency which authors should focus on for their publications?
\end{itemize}

Naturally, none of these questions is easy to answer due to the corresponding implications. In essence, we, as a community, are faced with a many-objective optimization problem where two of the objectives are sustainability and freedom of research. On the one hand, making any of the incentives for sustainability too strict and thus essentially designing and enforcing rules, for example, on the side of the funding agencies, bears the danger of limiting the freedom of research and therefore the potential progress of research quite strongly. On the other hand, completely ignoring sustainability is also not an option considering the global climate crisis, which impacts everyone and thus should also be addressed by everyone to a certain degree. In practice, the multi-objective problem actually has many more dimensions, such as social responsibility related aspects \citep{cheng2021socially}, FAIR data principles \citep{wilkinson2016fair}, and others. As a consequence, the community always has to strike a trade-off among all of these dimensions. Naturally, such a trade-off is hard to achieve and a rather continuous process, which is constantly shifting.

Due to the complexity of the matter, we believe that there is no clear answer to the questions raised in this section, and even more importantly, it is not clear who is responsible for answering them in particular. Nevertheless, we believe that every member of the research community can strive to achieve a trade-off that they deem good in their research. The most important thing is to stay open to new ideas and be aware of the different, perhaps not so obvious, aspects of one's research, such as sustainability or social responsibility.

\section{Prospects of AutoML}
\label{sec:prospects-of-automl}
While we have so far focused on how to make AutoML approaches themselves or research on AutoML greener, AutoML can in turn be used to make other systems or processes more sustainable as well \citep{van2021sustainable,tu2022AutoMLClimateChange}.
First, manually configuring machine learning algorithms is usually both time-consuming and inefficient. 
This is due to the fact that human intuition is usually a good heuristic for manageable problems, but rather unsuitable for traversing a high-dimensional space of possible candidate solutions, which in turn leads to a lower solution quality \citep{bergstra2011algorithms}.
AutoML can possibly remedy this situation, since corresponding systems typically try to traverse the search space as efficiently as possible, thereby also wasting less energy on uninformative candidate evaluations.

Second, AutoML gives a broader range of users access to ML technology, which in turn can be used to optimize other processes or (technical) systems, e.g., production facilities. For example, computationally complex simulations could be approximated by quick-to-evaluate ML models, so-called surrogate models, thus saving the energy to calculate the simulation \citep{reiner2021machine}. Data-driven surrogate models \citep{martins2021engineering} can be obtained using AutoML, which can further improve the sustainability aspect of surrogate models \citep{bliek2022survey}. In addition, other resources can be saved, such as spare parts, when AutoML is used for predictive maintenance tasks \citep{tornede2020automl,tornede2021coevolution}, for instance. While the actual task in predictive maintenance is to schedule maintenance cycles of plants more precisely so that the maintenance takes place as late as possible but still without any unplanned downtime or a breakdown of the plant, resources in terms of spare parts can also be saved. If parts of a plant are replaced too early, this not only costs money unnecessarily but also negatively affects the eco-balance, since the bottom line is that more spare parts are needed over time and therefore also have to be produced. Obviously, this production again requires energy and raw materials. Consequently, if AutoML enables more companies to reduce the usage of spare parts, AutoML can be leveraged to not only reduce the expenses of the company but also to save energy and physical resources.

Similarly, AutoML can also be used to find a model that predicts when renewable energy will be readily available in the energy distribution grid \citep{Wang19AutoMLElectricLoadForcasting}, and hence when energy extensive operations such as AutoML benchmarking should preferably be performed. This does not only help in making benchmarking itself more environmentally friendly but also in taking pressure off the distribution grid because too much energy in the distribution grid is as much a problem as too few. Overall, one can think of many such scenarios where data-driven models, possibly produced by AutoML, can contribute to sustainability.

\section{Conclusion}
In this paper, we proposed the idea of Green AutoML, a paradigm to make AutoML more efficient and environmentally friendly.
We have shown varying ways to determine the carbon emissions produced, motivated different methods that can be integrated with AutoML approaches to make the process more efficient, discussed existing methods as well as new ideas, and elaborated on strategies to reduce the required resources of the benchmark. Further work on empirical studies about the environmental footprint of AutoML systems could provide further insights into this topic. One could be a study about the behavior of different AutoML systems with varying \COe{} budget. Furthermore, we suggested giving detailed information about different aspects of the efficiency and environmental friendliness of the approach at the end of each published AutoML paper. In particular, we think that the environmental impact of a paper should be considered as a criterion in the review process.  
On the other side, failed attempts and negative results should also be published to avoid duplicated work on the same ideas. In general, appropriate research incentives can push the research in the direction of Green AutoML due to special issues and special tracks of journals and conferences. Furthermore, the funding agencies should force the researchers to work in the direction of Green AutoML through corresponding calls for projects or acceptance criteria for projects, and allow also \COe{} compensation for experimental analysis. Additionally, one could question if it is reasonable to publish a paper at a conference hosted on the other side of the world if an on-site presentation and thus, a long and \COe{} expensive flight is required. Regardless of the aforementioned aspects, it is the community that has to take action, sooner rather than later.

\section*{Acknowledgements}
This work was partially supported by the German Federal Ministry of Education and Research (ITS.ML project no.\ 01IS18041D) and the German Research Foundation (DFG) within the Collaborative Research Center ``On-The-Fly Computing'' (SFB 901/3 project no.\ 160364472).

We also thank the COSEAL community for their feedback on a poster based on this work at the COSEAL workshop in 2021 and the AutoML Fall School 2021 organizers for the opportunity to highlight the topic of Green AutoML at the panel discussion. Moreover, we would like to thank the panelists for their remarks and ideas on the topic, which have been incorporated into this work.

Lastly, we thank the anonymous reviewers for their feedback and pointers to interesting related areas of research and papers, which helped to make this a better paper.

\printbibliography

\end{document}